%% file: 3DV_joint_reconstruction.tex
\documentclass[10pt,twocolumn,letterpaper]{article}

\usepackage{3dv}
\usepackage{times}
\usepackage{epsfig}
\usepackage{graphicx}
\usepackage{amsmath}
\usepackage{amssymb}

\usepackage{algorithm}
\usepackage{algpseudocode}
\usepackage{algorithm}

\usepackage[pagebackref=true,breaklinks=true,letterpaper=true,colorlinks,bookmarks=false]{hyperref}

\renewcommand{\vec}[1]{\mathbf{#1}}
\newcommand{\mat}[1]{\mathbf{#1}}

\threedvfinalcopy 

\def\threedvPaperID{164} 

\ifthreedvfinal\fi
\begin{document}

\title{
Joint 3D Reconstruction of a Static Scene and Moving Objects
}

\author{Sergio Caccamo\\
KTH Royal Institute of Technology\\
Stockholm, Sweden\\
{\tt\small caccamo@kth.se}
\and
Esra Ataer-Cansizoglu and Yuichi Taguchi\\
Mitsubishi Electric Research Labs\\
Cambridge, MA, USA\\
{\tt\small \{cansizoglu,taguchi\}@merl.com}
}

\maketitle
\thispagestyle{empty}

\begin{abstract}

We present a technique for simultaneous 3D reconstruction of static regions and rigidly moving objects in a scene. An RGB-D frame is represented as a collection of features, which are points and planes. We classify the features into static and dynamic regions and grow separate maps, static and object maps, for each of them. To robustly classify the features in each frame, we fuse multiple RANSAC-based registration results obtained by registering different groups of the features to different maps, including (1) all the features to the static map, (2) all the features to each object map, and (3) subsets of the features, each forming a segment, to each object map. This multi-group registration approach is designed to overcome the following challenges: scenes can be dominated by static regions, making object tracking more difficult; and moving object might have larger pose variation between frames compared to the static regions. We show qualitative results from indoor scenes with objects in various shapes. The technique enables on-the-fly object model generation to be used for robotic manipulation. 

\end{abstract}

%

\input{introduction.tex}

\input{relatedworks.tex}
\input{methodology.tex}

\input{experiments.tex}

\input{conclusions.tex}

\textbf{Acknowledgments:}
This work was done at and supported by Mitsubishi Electric Research Laboratories. We thank the anonymous reviewers and Wim Abbeloos for their helpful comments.

\addtolength{\textheight}{-2.5cm}   

{\small
\bibliographystyle{ieee}
\bibliography{SLAM}
}

\end{document}

%% file: introduction.tex
\section{Introduction}
\label{sec:intro}

Scene understanding and navigation are crucial for autonomous agents to localize themselves with respect to a reconstructed map and interact with the surrounding environment. 3D object modeling and localization lie at the core of robot manipulation. Conventional simultaneous localization and mapping (SLAM) systems are successful for representing the environment, when the scene is static. Yet, in the case of a dynamic scene, large moving objects can degrade the localization and mapping accuracy. On the contrary, object motion can provide useful object information acquired from various viewpoints. 

\begin{figure}[t]
  \center
     \includegraphics[width=0.8\columnwidth]{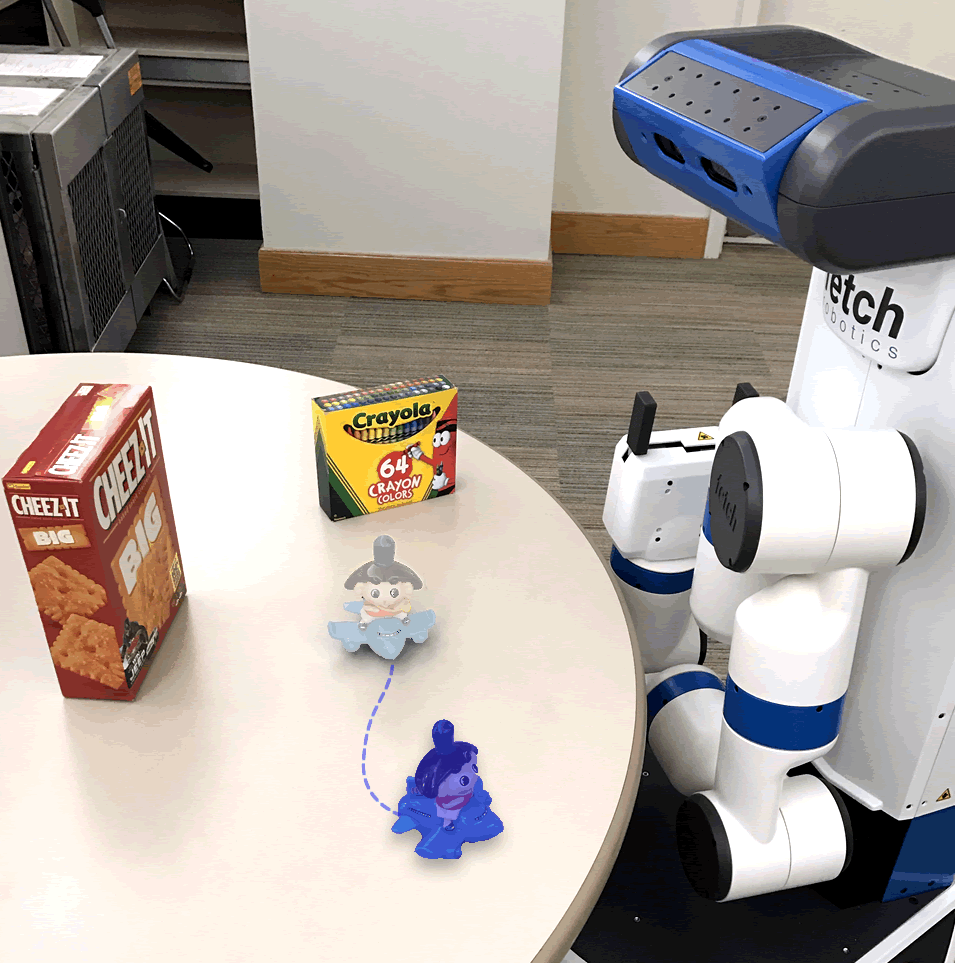}
  \caption{Illustrative representation of the system. The mobile robot used in the experiments detects a moving object and generates an object map separately from a static map corresponding to the static environment.}
  \label{fig:concept}
\end{figure}

\begin{figure*}[ht]
\centering
\includegraphics[width=0.85\textwidth]{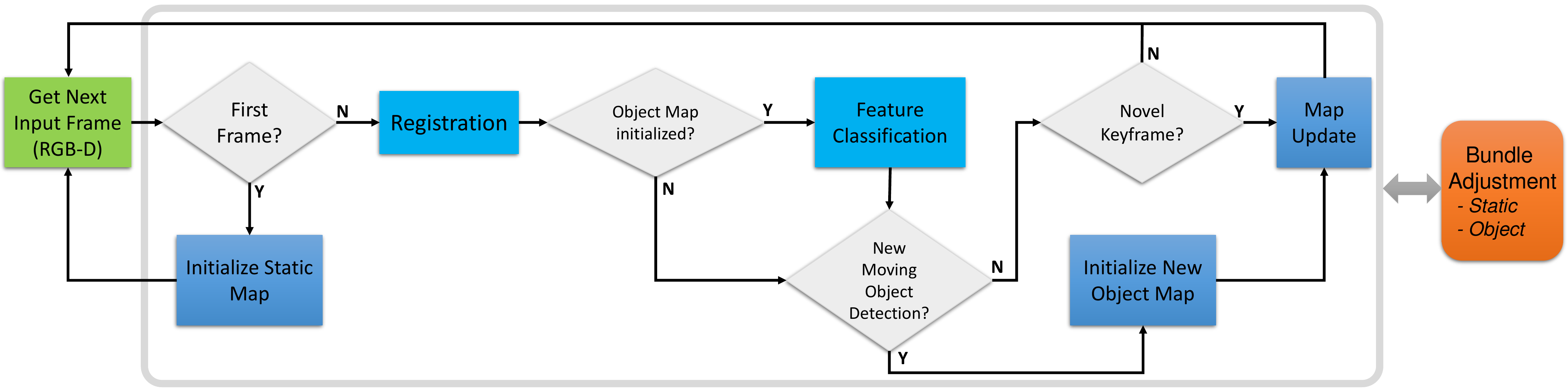}
\caption{System overview: Our method first initializes the static map with the first input frame and captures another RGB-D frame. At the next step, registration module performs a multi-group registration between the current input frame and each of the existing maps. If no object maps are initialized yet, moving object detection module finds the regions that belong to the object. If there are already existing object maps, then we first perform feature classification and split the measurements associated to the existing object maps. For the rest of the measurements, we run moving object detection in order to find if there are new objects in the scene. Depending on the estimated pose of the frame with respect to each map, the frame is added as novel keyframe to the respective map. Bundle adjustment procedure runs asynchronously with SLAM.}
\label{fig:system}
\end{figure*}

This paper presents a technique for simultaneous reconstruction of static regions and rigidly moving objects in a scene. While each of the camera localization and moving object tracking is already a challenging problem, addressing both of them simultaneously creates a chicken-and-egg problem: It is easy to map a scene when object motion is known and object regions can be removed from input frames beforehand. On the other hand, it is easy to detect moving object, when the camera pose is known. The presented method creates independent maps for static scene and moving objects, by tackling both problems following a multi-group registration scheme.
 
We first start with a single map and localize each frame with respect to this map, referred to as a static map. A frame is represented as a collection of segments, where each segment contains a group of features extracted from the frame. A moving object is detected as a set of segments that has high outlier ratio after frame localization with respect to the static map. Once we detect the features that fall inside dynamic segment measurements, we initialize a new map to represent the rigidly moving object, referred to as an object map. In the following observations, each frame is registered with respect to both the static and object maps. We distinguish the features belonging to the objects and static region based on the inliers resulting from these registrations.

Our main contribution is the accurate discrimination of features coming from dynamic and static regions following a multi-group geometric verification approach. Following the approach of~\cite{Cansizoglu16IROS}, we use feature grouping for object representation. In our SLAM framework, the keyframes are treated as a collection of features and objects are seen as a collection of segments, that are subset of features from keyframe. Our multi-group registration scheme considers registration of all features and various subsets of features of the frame against each map. First, we register all measurements against the static map and the object map. If there are sufficient features coming from both static and moving objects, this frame-based registration will succeed for both maps. However, if there is a dominating motion pattern in the frame, then localization of small moving objects can be missed. Thus, we also carry out a segment-based registration procedure, where we register the features falling inside a segment against the object map. To perform robust feature classification, we fuse these registration results obtained from multiple geometric verifications. 

Although the technique in~\cite{Cansizoglu16IROS} also deals with object tracking on a SLAM framework, there are major differences with this study. First, the method in~\cite{Cansizoglu16IROS} only involves localization of static objects inside a SLAM system and it does not address the problem of forming multiple independent maps. On the other hand, in this study we further tackle the problem of classifying features into static and dynamic maps after localizing objects. Distinguishing features and building disjoint maps is challenging, as any contamination from one set to the other will severely affect the localization afterwards. 
Second, the method of~\cite{Cansizoglu16IROS} will not work for moving objects in a sequence, as localization and bundle adjustment strongly relies on the static scene assumption. In this paper, we handle moving objects and do not have any assumption about the motion of the object (i.e., smooth or abrupt). The object motion is utilized as a means of 3D object model construction, as the  motion provides a rich viewpoint variation of the object. Third, we provide a simultaneous navigation and object modeling system, while a separate object scanning step is required for object tracking in~\cite{Cansizoglu16IROS}.

An important advantage of our method is on-the-fly generation of object models, while mapping static environment at the same time. Just like a kid learning to model and manipulate objects by watching others, our method learns both object model and static scene map at the same time based on the motion of the object. An example use of the presented technique is simultaneous robot navigation and object modeling as seen in Figure~\ref{fig:concept}.

\subsection{Contributions}
We summarize the contributions of our work as follows:
\begin{itemize}
\item a multi-group registration scheme used to distinguish features coming from regions with different motion patterns, i.e., static region and moving object.
\item simultaneous reconstruction of static and object maps, yielding on-the-fly object model generation.
\item an automated technique to detect moving objects in a SLAM framework.
\end{itemize}

%% file: relatedworks.tex
\subsection{Related Work}
\label{sec:relatedWork}

Object SLAM aims to detect and track static objects occurring multiple times in a sequence of frames and use this information for building more accurate maps~\cite{Dharmasiri2016,Civera2011,Ma2014,SalasMoreno13CVPR,Fioraio2013}. Although it is widely studied in the scope of stationary objects, there are few studies on moving object tracking in an RGB-D SLAM framework~\cite{Cadena2016}. 

Existing work on dynamic object tracking either focuses on detecting the moving object to remove it from the static map~\cite{Keller2013, Tan2013}, or generating a model of the moving object by ignoring reconstruction of static environment~\cite{Jiang2016, Mustafa2015, Yuheng2013, Shin2013, Ambrus2017,  Dame2013}. Keller et al.~\cite{Keller2013} solved dynamic object detection and camera localization in alternating steps in order to remove moving objects from the reconstructed map. They have a dense SLAM system, which makes it computationally demanding compared to sparse feature-based systems. Similarly,~\cite{Mustafa2015, Yuheng2013, Dame2013} used a dense point cloud representation in order to generate 3D reconstruction of objects without modeling the static scene. Jiang et al.~\cite{Jiang2016} presented an object tracking method based on motion segmentation, hence making the system unable to operate online. Shin et al.~\cite{Shin2013} developed a framework for 3D reconstruction of multiple rigid objects from dynamic scenes. Their framework provides reliable and accurate correspondences of the same object among unordered and wide-baseline views, providing reconstruction of the object only. 

Choudhary et al.~\cite{Choudhary2014} presented a method to create 3D models of static objects in the scene while performing object SLAM at the same time. Their method depends on segment associations and cannot handle objects that rigidly move with respect to the static scene. Finman et al.~\cite{Finman2013} created 3D models of objects by using differences between RGB-D maps, where they focused on re-identification of objects rather than generating complete 3D models. In a similar way, F{\"a}ulhammer et al.~\cite{Faulhammer2017} used  segmentation of dense point cloud to generate 3D model of an object in an indoor environment, while mobile robot patrols a set of points.

The closest work to our technique was proposed by Wang et al.~\cite{Wang2007}, where they use a monocular camera and track the camera and dynamic object separately following a probabilistic approach. However, rather than constructing a complete 3D model of the object, their method only keeps track of the object locations while performing SLAM in the static environment. Moreover, their system can have difficulties when the object does not have any smooth motion (i.e., manipulated by a human). Recently Ataer-Cansizoglu and Taguchi~\cite{Cansizoglu16IROS} presented a technique to track objects in RGB-D SLAM following a hierarchical grouping approach. An important limitation of their algorithm is the inability to handle moving objects. Furthermore, their technique includes a separate object scanning step that generates the 3D model of the object, which is used later for object tracking and localization. On the other hand, this work focuses on on-the-fly object model generation and tracking of rigidly moving objects.

%

%% file: methodology.tex
\section{Methodology}
\label{sec:Method}

We build our framework on the pinpoint SLAM system~\cite{Taguchi13ICRA,Cansizoglu16ICRA}, which localizes each frame with respect to a growing map using 3D planes, 3D points, and 2D points as primitives. The use of 2D measurements enables to exploit information in regions where the depth is not available (e.g., too close or far from the sensor). In this paper, our segments include 3D points and 3D planes (but not 2D points) as features similar to~\cite{Cansizoglu16IROS}, while the registration procedure exploits all the 2D points, 3D points, and 3D planes as features. We use the standard terminology of \textit{measurements} and \textit{landmarks}. Namely, the system extracts measurements from each input frame and generates/updates landmarks in a map. We use SIFT~\cite{Lowe04IJCV} detector and descriptor for generating 2D and 3D point measurements, while 3D plane measurements are extracted using the method of~\cite{Feng14ICRA}.

Figure~\ref{fig:system} shows an overview of our system. Our method consists of three modules. Dynamic object detection module finds a group of features coming from a moving object in order to initialize a map for the object, referred to as an object map. Registration module involves localization of the input frame with respect to each of the existing maps, including a static map corresponding to the static environment and a set of object maps. We perform a multi-stage registration to ensure accurate pose estimates, since distinguishing features into static and dynamic regions relies on the registration output. Finally, feature classification module divides the features into groups by detecting which map they are associated to based on the localization results. Map update and bundle adjustment procedures are carried out for enlarging the maps and refining the keyframe poses respectively.

\begin{figure}[t]
  \center
     \includegraphics[width=\columnwidth]{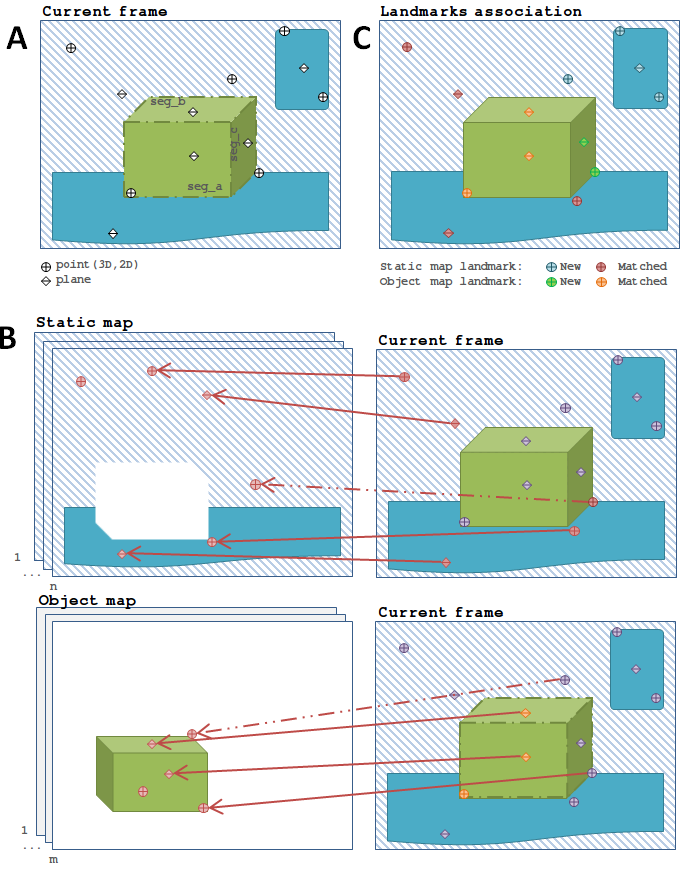}
  \caption{Landmarks association: (A) Measurements are extracted from the frame where points and planes are grouped into segments. (B) Registration is performed between all extracted measurements and the landmarks in the static map and object map. A further segment-based registration (as in Section~\ref{sec:localization}) is performed on the object map that helps remove erroneously matched measurements (dashed lines) as described in Section~\ref{sec:classificationland}. (C) The landmarks are assigned to the object map and static map.}
  \label{fig:landmarksassoc}
\end{figure}

\subsection{Moving Object Detection}
Following a feature grouping approach, we generate segments in the frame after feature extraction. We first make use of plane extraction output and initiate a segment from each plane, i.e., each segment contains the extracted plane and all point features that fall inside the plane. Second, we carry out a depth-based segmentation algorithm to generate segments from the rest of the point cloud after plane extraction.

In this work, the first input frame initializes the static map\footnote{The map that is initialized with the first frame corresponds to the regions of the scene with the dominant motion pattern observed at the initial keyframes. Thus, without loss of generality, we use the phrase ``static map'' with the assumption that dominant motion pattern comes from the static region.}. Next, we register each following frame with respect to the static map, which results in inlier and outlier measurements. When the static regions dominate the frame, this procedure will result in the pose of the static frame. Thus, in order to detect segments that belong to the moving object, we consider the number of outliers per segment. If a segment has a large ratio of outlier measurements, then it is considered as an object region. All the features that fall inside the segment are used to initialize a new object map for the detected segments.

Dynamic object detection is executed for each frame, enabling initialization of multiple object maps. Hence, our system is capable of growing multiple object maps referring to different moving objects.

\subsection{Registration}
\label{sec:localization}

Each frame consists of sets of features coming from the moving object and the static region. We employ a multi-group registration scheme to verify the groups of features associated to different maps. We first perform registration between all the features of the frame and each existing map independently. Since objects might be small, this frame-based registration might fail for the object maps. Thus, we proceed with a segment-based registration that aims to register groups of features that are represented by segments against each object map.
After these registrations, we come up with multiple pose estimations for the input frame with respect to the maps. If both registrations succeeded, we use the result of the segment-based registration as the pose estimate since it achieves a more robust correspondence search due to a smaller number of measurements in segments. We also fuse inlier outputs from both registrations by prioritizing segment-based registration output.

\subsubsection{Frame-based Registration}
We match all features extracted from the frame with all features that come from the last $N$ keyframes of the target map.  Let $\vec{p}_m$ be a measurement extracted from the input frame and $\vec{p}_l\in L$ be the corresponding landmark of the target map according to feature matching with the set of landmarks $L$. Let us denote the set of measurements of frame $i$ as $F_i$. By exploiting measurement-landmark matches in a RANSAC framework, we estimate frame-based pose $\mat{\hat{T}}_i$ by solving the following problem:
\begin{equation}
\mat{\hat{T}}_i = \underset{\mat{T}_i}{argmin} \sum_{\vec{p}_m \in I_i} d( \mat{T}_i(\vec{p}_m), \vec{p}_l ).
\label{eqn_frame}
\end{equation}
Here $\mat{T}(\vec{p}_m)$ indicates the transformation of measurement $\vec{p}_m$ by pose $\mat{T}$, and $d(\cdot, \cdot)$ denotes distances between features, which are defined for 3D-to-3D point correspondences, 2D-to-3D point correspondences, and 3D-to-3D plane correspondences as in~\cite{Cansizoglu16ICRA}. $I_i$ is the set of inlier measurements detected as 
\begin{equation}
I_i = \{ \vec{p}_m \in F_i | \exists \vec{p}_l \in L \textrm{ s.t. }  d(\mat{\hat{T}}_i(\vec{p}_m),  \vec{p}_l) < \sigma \},
\end{equation}
where $\sigma$ is an inlier threshold.
Note that, for the static map, since the camera moves smoothly, restricting the features to the last $N$ keyframes provides faster correspondence search. On the other hand, if the object motion is abrupt, it is likely that frame-based registration can fail. Thus, we proceed with segment-based registration for localizing the frame with the object maps.

\subsubsection{Segment-based Registration}
As proposed in~\cite{Cansizoglu16IROS}, we detect and track objects by performing a segment-based registration with respect to the object maps. An object map is represented as a collection of segments that are registered with each other. For each segment in the input frame, we perform VLAD-based appearance similarity search~\cite{Jegou12PAMI} followed by RANSAC registration to register the segment with respect to an object map. 
 Let us denote the set of measurements of segment  $j$ of frame $i$ as $S_{i,j} \subset F_i$, and the set of landmarks of the matching segment as $L_j \subset L$. Similar to frame-based registration, we carry out feature matching between $S_{i,j}$ and $L_j$ and solve the following optimization problem through RANSAC:
\begin{equation}
\mat{\hat{T}}'_{i,j} = \underset{\mat{T}_{i,j}}{argmin} \sum_{\vec{p}_m \in I'_{i,j}} d(  \mat{T}_{i,j}(\vec{p}_m), \vec{p}_l). 
\label{eqn_segment}
\end{equation}
Here $\mat{\hat{T}}'_{i,j}$ is the estimated object pose and $I'_{i,j}$ is the set of inlier measurements detected as
\begin{equation}
I'_{i,j} = \{ \vec{p}_m \in S_{i,j} | \exists \vec{p}_l \in L_j \textrm{ s.t. }  d(\mat{\hat{T}}'_{i,j}(\vec{p}_m), \vec{p}_l) < \sigma \}.
\end{equation}
Since this pose is based on segment-to-segment registration, we proceed with a final refinement carrying out a prediction-based registration between all the measurements of the frame and the landmarks of the object map similar to~\cite{Cansizoglu16IROS}. In other words, we use the result of equation~\eqref{eqn_segment} as the predicted pose, and perform feature matching between the frame and the map based on that. Then, using these matches we perform  RANSAC that minimizes the error between the measurements of the frame and the landmarks of the map as indicated in equation~\eqref{eqn_frame}, obtaining the refined object pose $\mat{\hat{T}}_{i,j}$. After refinement the inliers of segment $S_{i,j}$ are
\begin{equation}
I_{i,j} = \{ \vec{p}_m \in F_i | \exists \vec{p}_l \in L \textrm{ s.t. }  d(\mat{\hat{T}}_{i,j}(\vec{p}_m), \vec{p}_l) < \sigma \}.
\end{equation}
Note that since final refinement is performed between all measurements of the frame and the map, there might be inliers that are outside of segment $S_{i,j}$ as indicated in the above equation. This way, we can handle the object features that do not belong to any segment and/or have invalid depth values, for example the features in small object regions that are missed during segmentation due to depth discontinuity or invalid depth values.

This step outputs the pose of the object in the current frame with respect to the object map and the matching segments of the frame along with the associations between the measurements and the landmarks of the object map. In the following step, we proceed with a classification method to distinguish features with different motion patterns using the registration output.

\subsection{Classification of Features into Regions}
\label{sec:classificationland}

The multi-group registration provides us pose estimates of the input frame with respect to the static map and object maps along with the associations between measurements and the landmarks. The segment-based registration also outputs the segments that are successfully matched with a segment in an object map.

Since the objects are smaller compared to the static scene and motion of the static region dominates the scene, we prioritize object maps while classifying the measurements. Thus, if a measurement falls inside a segment that is matched with a segment of an object map, then the measurement is considered as associated to the object. Otherwise, we investigate whether any of the registrations found the measurement as an inlier. If the measurement is found as an inlier in the object map registration, then it is considered as object measurement. Otherwise, the measurement is considered as belonging to the static scene. This means that at the end of this process the measurements extracted from the novel frame are binary associated to the two maps as shown in Figure \ref{fig:landmarksassoc}. 

The steps of the method are summarized in Algorithm~\ref{alg:fusion}. In lines 1--2, $M^{static}$ and $M^{object}$ are initialized to empty sets, that keep measurements associated to static and object maps respectively. Frame-based registration is carried out with respect to both maps in lines 3--6, followed by segment-based registration in lines 7--13. Feature classification updates $M^{static}$ and $M^{object}$ in lines 14--24 and the maps are updated in lines 25--26. Note that map update does not happen if none of the registrations succeeds for the map.

\begin{algorithm}[t]
\caption{Algorithm for updating maps given frame measurements $F_i$, measurements of segments $S_{i,1}, S_{i,2},\ldots,S_{i,n}$, and the set of landmarks of static and object maps,  $L^{static}$ and $L^{object}$. }\label{alg:fusion}
\begin{algorithmic}[1]
\small
\State{$M^{static} \gets \varnothing $} \Comment{measurements associated to static map}
\State{$M^{object} \gets \varnothing $} \Comment{measurements associated to object map}

\State{\texttt{Match features between $F_i$  and $L^{static}$}}
\State{\texttt{Compute $\mat{\hat{T}}_i^{static}$ and ${I}_i^{static}$ by eqn.~\eqref{eqn_frame}}}

\State{\texttt{Match features between $F_i$  and $L^{object}$}}
\State{\texttt{Compute $\mat{\hat{T}}_i^{object}$ and ${I}_i^{object}$ by eqn.~\eqref{eqn_frame}}}

\For{$j=1,\ldots,n$}
	\State{\texttt{Match features between $S_{i,j}$ and $L_j$}}
	\State{\texttt{Compute $\mat{\hat{T}}'_{i,j}$ and $I'_{i,j}$ by eqn.~\eqref{eqn_segment}}}
	\State{\texttt{Match features between $F_i$  and $L^{object}$ based on $\mat{\hat{T}}'_{i,j}$}}
	\State{\texttt{Compute $\mat{\hat{T}}_{i,j}$ and $I_{i,j}$ by eqn.~\eqref{eqn_frame}}}
	\State{\texttt{Report $S_{i,j}$ as matching segment if RANSAC succeeds}}
\EndFor

\For{$\forall \vec{p}_m \in F_i$}
	\If {\texttt{$\vec{p}_m$ is inside a matching segment}}
  		 \State { $M^{object}\gets M^{object} \cup \{\vec{p}_m\}$}
	\Else
    		\If {\texttt{$\vec{p}_m \in I_i^{object}$ or $\exists S_{i,j} | \vec{p}_m \in I_{i,j}$}}
       			\State { $M^{object}\gets M^{object} \cup \{\vec{p}_m\}$}
		\Else
			\State { $M^{static}\gets M^{static} \cup \{\vec{p}_m\}$}
    		\EndIf
	\EndIf
\EndFor
\State{\texttt{Update $L^{static}$ with $M^{static}$}}
\State{\texttt{Update $L^{object}$ with $M^{object}$}}

\end{algorithmic}
\end{algorithm}

\begin{figure*} [t]
  \center
     \includegraphics[width=\textwidth]{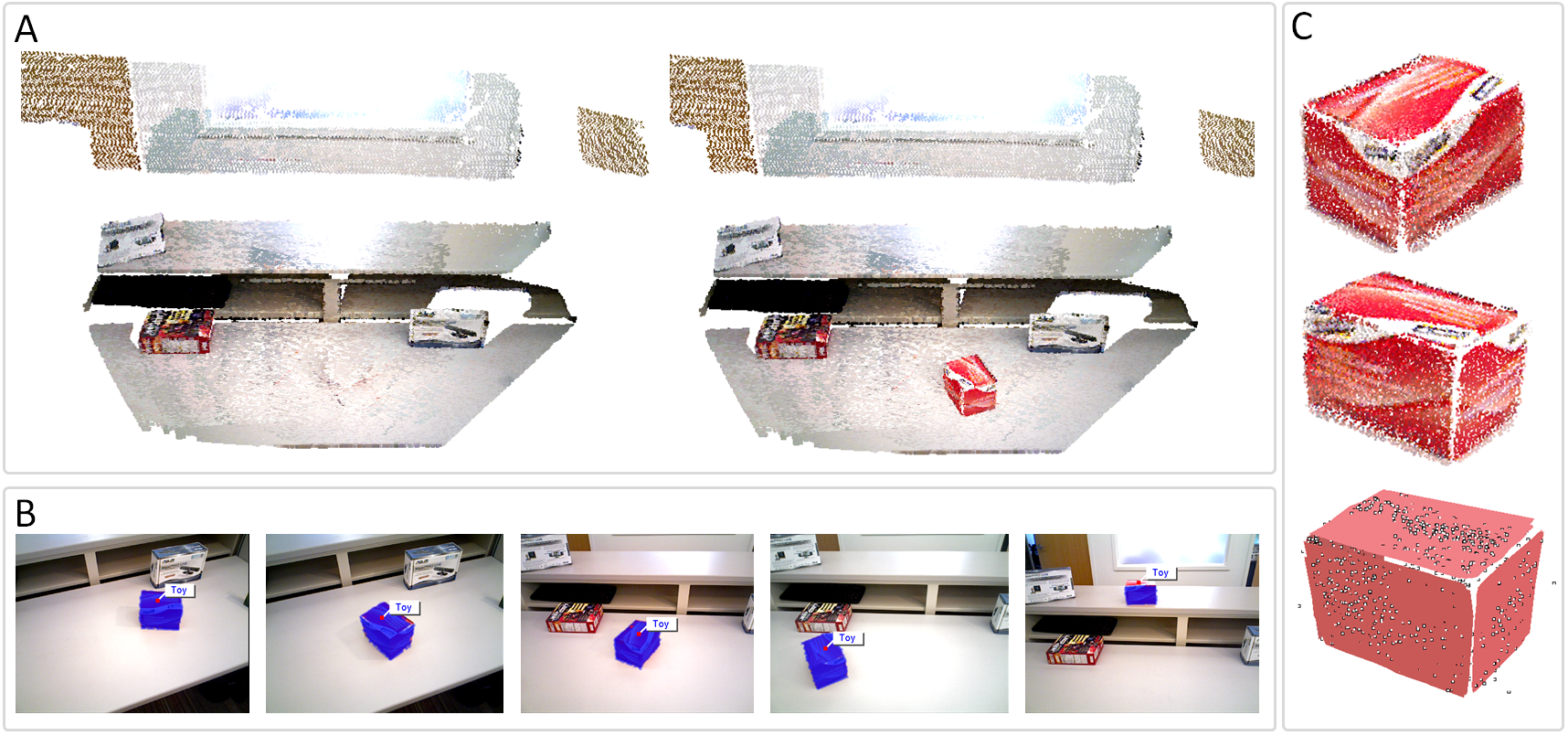}
  \caption{Reconstruction results on example scene 1: (A) 3D reconstructed static map (left) and object map overlaid on the static map based on the initial pose of the object (right), (B) example keyframes from the sequence where the leftmost frame is the first keyframe of the object map after automatic moving object detection, (C) reconstructed 3D map of the moving object in various viewpoints (top and middle) and point landmarks overlaid on plane landmarks with white circles (bottom). }
  \label{fig:pic_experiment1}
\end{figure*}

\subsection{Map Update and Bundle Adjustment}

After the registration, we know which group of features are associated to static regions or the objects. We also have a pose estimation for each group of features with respect to the map they are associated to. For each map, if the estimated pose is different from the poses of existing keyframes of the map, then we initialize a new keyframe with the respective set of features and add the keyframe to the map. If registration fails for one of the maps, then the map is not updated with any information from that frame.

A bundle adjustment procedure runs asynchronously with the SLAM for each map, minimizing the registration error with respect to all the keyframe poses and landmark locations. Note that since the motion of the sensor and the motion of the objects are independent from each other, we do not utilize any constraints based on object correspondences in the bundle adjustment contrary to the approach in~\cite{Cansizoglu16IROS}.

%% file: experiments.tex
\section{Experiments and Results}
\label{sec:Experiments}

We evaluated our method on different indoor scenes recorded from either a hand-held RGB-D camera (ASUS Xtion) or a mobile Fetch robot as shown in Figure \ref{fig:concept}.
The system was implemented in C++ on the Robot Operating System (ROS), and used images and depth maps at resolution of $640 \times 480$ pixels. We used $0.4$ as the RANSAC inlier ratio and set $\sigma = \max(1, 3 \sigma_{Z})$ in cm where $\sigma_{Z}$ is the depth-dependent measurement error~\cite{Khoshelham2012Accuracy} for deciding whether measurement and landmark associations are inliers. We did not proceed with RANSAC and reported localization failure if there were less then 10 feature matches. Bundle adjustment was performed using the Ceres Solver \cite{CeresSolver}. This online SLAM system runs at $\sim 3.5$ frames per second on CPU. The experiments described below aim to test the capability of our system on (i) simultaneously and independently reconstructing the static scene and rigidly moving objects and (ii) detecting and tracking moving objects \footnote{Video of the experiments available at \url{https://youtu.be/goflUxzG2VI}}.

\subsection{Experimental Scenarios}
\label{sec:expscenes}

\textbf{Scene 1} (Figure~\ref{fig:pic_experiment1}): For the first experiment we used a discrete set of RGB-D images showing different objects placed on a desk captured from different viewpoints. The red box was the only moving object in the scene. As soon as the red box moved (first frame in Figure \ref{fig:pic_experiment1}) the system initialized the static and object maps and started tracking the object. Figure \ref{fig:pic_experiment1}(B) shows some of the keyframes stored in the object map along with the position of the red box. The superimposed blue mask indicates the frame segments associated to the object map (i.e., sets of features fed to the object map). Figure \ref{fig:pic_experiment1}(A) shows the reconstructed static map (left) which contained $10$ keyframes, $2270$ point landmarks, and $17$ plane landmarks, as well as the combined object and static map (right). The object model is placed on the initial detected position. Figure \ref{fig:pic_experiment1}(C) shows the reconstructed object model and the object map ($11$ keyframes) having $813$ point landmarks and $4$ plane landmarks. Notice that the system is able to decouple the two maps and does not require smooth object motion. 

\begin{figure}[t]
  \center
     \includegraphics[width=\columnwidth]{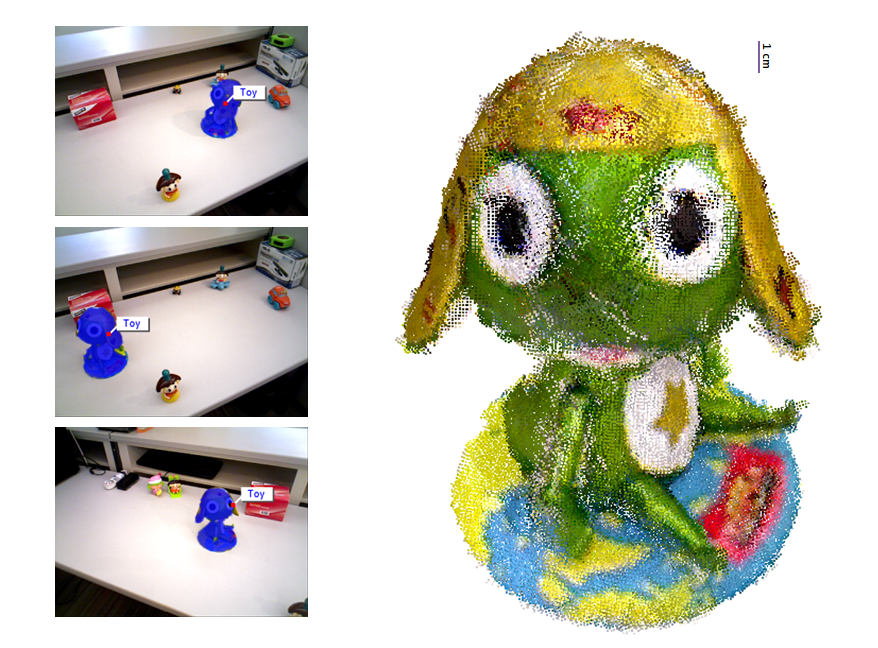}
  \caption{Reconstruction results on example scene 2, where we model an object which consists of mostly non-planar segments: example keyframes from the sequence (left), reconstructed point cloud (right).}
  \label{fig:pic_experiment2}
\end{figure}

\textbf{Scene 2} (Figure~\ref{fig:pic_experiment2}):  In the second experiment we used the robot in Figure \ref{fig:concept} to look at various objects on a desk. When the green toy moved the system initialized the object map. In this experiment, we show that the proposed method is able to model an object consisting of mostly non-planar regions. The object map included $812$ point landmarks and $3$ plane landmarks whereas the static map had $3038$ point landmarks and $11$ plane landmarks.

\begin{figure*} [t]
  \center
     \includegraphics[width=\textwidth]{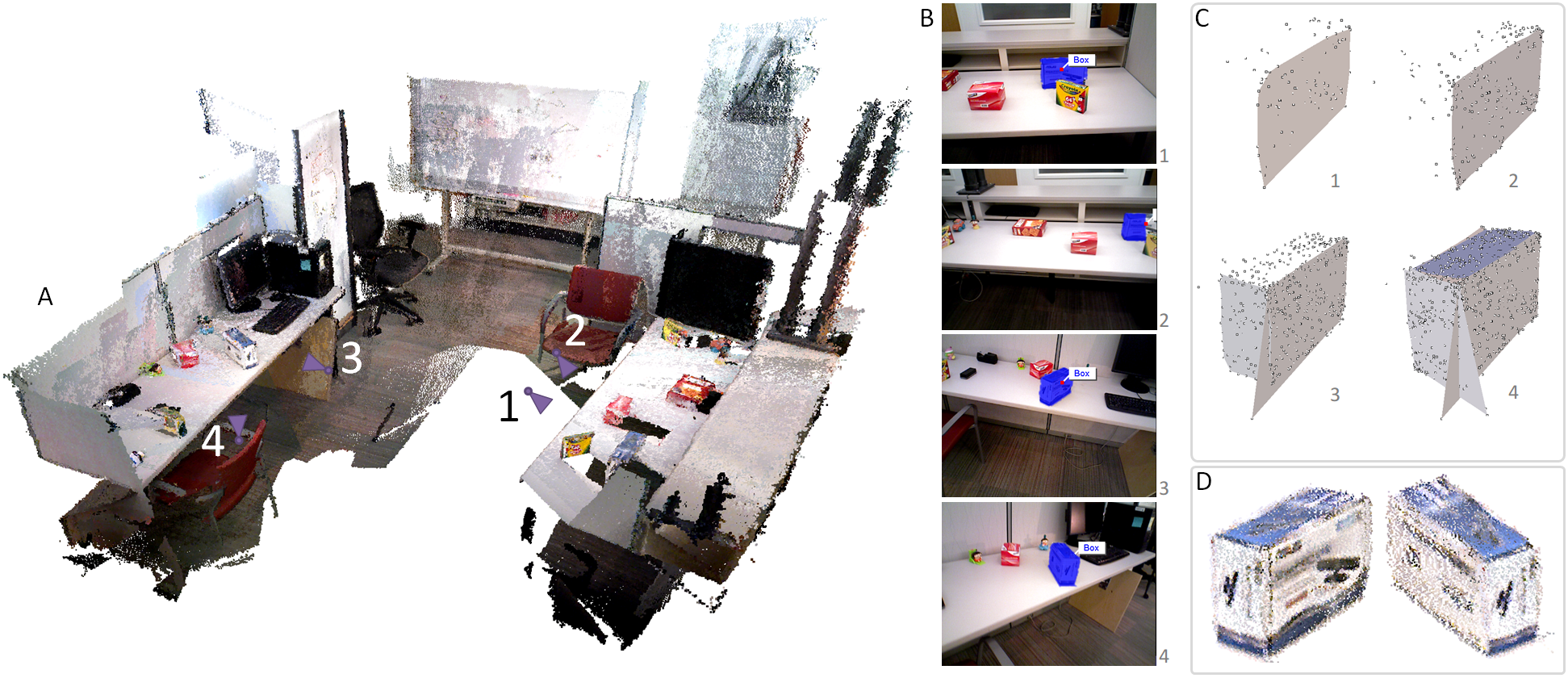}
  \caption{Reconstruction results on example scene 3: (A) reconstructed point cloud for static and object maps along with the 4 camera locations, where keyframes were added to both maps, (B) keyframes of the camera locations shown in (A), where blue color indicates the set of segments added to the object map, (C) point landmarks overlaid on the plane landmarks of the object map when respective keyframes were added, (D) point cloud visualization of the reconstructed object map from two different viewpoints. Notice that although the object map was partially occluded in the initial keyframe, the final reconstructed model was gradually completed using measurements from other keyframes.}
  \label{fig:pic_experiment3}
\end{figure*}

\begin{figure} [t]
  \center
     \includegraphics[width=\columnwidth]{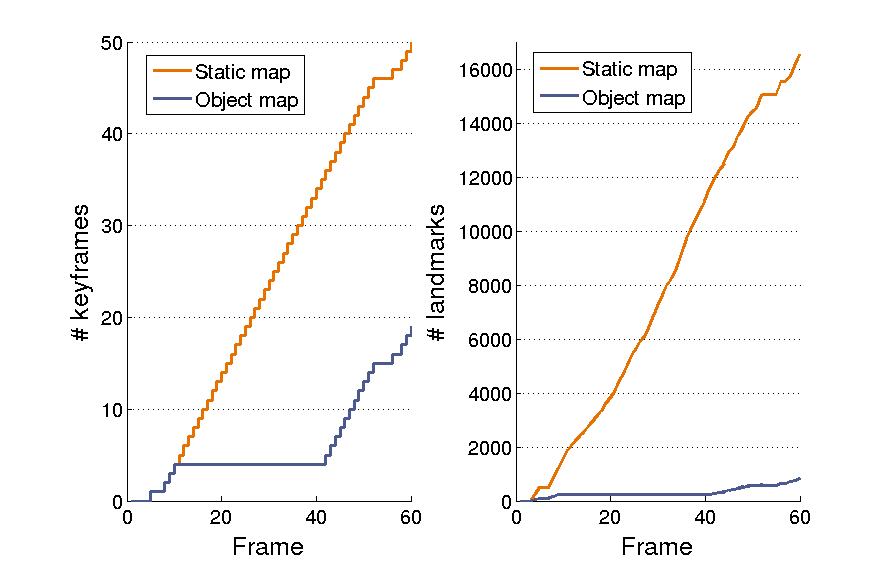}
  \caption{Plot of number of keyframes (left) and number of landmarks with respect to frame indices for static (red) and object (blue) map of experiment scene 3. The static map grows larger, while the object map only enlarges when the object is visible.}
  \label{fig:chart_experiment3}
\end{figure}

\textbf{Scene 3} (Figure~\ref{fig:pic_experiment3}): In the third experiment we used a hand-held RGB-D camera on an indoor office scene. The scene contained two instances of the target object (white-blue box) placed on different locations. Figure \ref{fig:pic_experiment3}(B) shows the keyframes added to both static and object maps during the whole experiment. The user started the experiment by pointing the camera at the white-blue box which was initially partially occluded as seen in position 1 of Figure~\ref{fig:pic_experiment3}(A). In this experiment, we manually specified the segments corresponding to the object in the first frame to initialize an object map since the object was stationary. The user then moved the camera away from the box, focusing on the rest of the office (frame $10$, the position 2). The system lost track of the object and stopped adding new keyframes to the object map. After a brief exploration ($32$ frames), the user pointed the camera at the second instance of the white-blue box (frame $42$, the position 3).
The system relocalized the object, generated a new keyframe based on feature classification as described in Section~\ref{sec:Method}, and started adding new keyframes to the object map. The number of landmarks increased as shown in Figure \ref{fig:pic_experiment3}(C), where we display point landmarks overlaid on the plane landmarks of the object map when respective frames were gradually added. This is possible because the two maps were always decoupled and the system always performed independent global registration of the current frame with respect to the two maps. Thus the registration failure of the frames $10 \to 42$ against the object map was not a problem for the system to relocalize the object again. The failure did not stop the growth of the static map in those frames since static map localization did not lose track. The user then moved the camera around the box (the position 4 in Figure \ref{fig:pic_experiment3}) and new plane and point landmarks were added to the object map as seen in Figure \ref{fig:pic_experiment3}(C), which shows the evolution of the object map on the described keyframes. Our plane extraction algorithm fitted closeby points into the plane boundaries resulting in the leaking representation of Figure \ref{fig:pic_experiment3}(C3-4). This, however, does not affect plane registration, which only considers plane equations. The reconstructed box model is displayed in Figure \ref{fig:pic_experiment3}(D) which was generated using all the estimated keyframe poses contained in the object map. Similarly, the combination of the static map (office) and the two box instances displayed in Figure~\ref{fig:pic_experiment3}(A) was generated using all the estimated keyframe poses contained in the static and object maps. Figure~\ref{fig:chart_experiment3} shows the chart for number of landmarks and number of keyframes with respect to the input frame indices. As can be seen, the static map grows larger whereas the object map only grows when the object is visible.

%% file: conclusions.tex
\section{Conclusion and Discussion}
\label{sec:conclusions}

We presented a novel real-time SLAM system that jointly reconstructs a static scene and moving objects from either continuous or discrete observations (i.e. no smooth object motion required). The system automatically detects and models a moving object from the static scene and creates two independent maps. It extracts 3D points, 2D points, and 3D planes from RGB-D data and splits them into disjoint measurement sets that are independently used by the two maps for stable registration. The use of a sparse feature-based representation allows continuous and independent optimization for the two maps even on CPU.  Thus, a mobile robot can reliably model an object during exploration and then use the reconstructed model for manipulation tasks. Note that we avoided modeling hand-object interactions in this work, as the focus was simultaneous reconstruction of multiple maps from various motion patterns. However, the use of the presented system on a mobile robot platform is possible by disabling SLAM system every time robot hand enters into the view to interact with the object and enabling it back when the hand is not visible.

Our moving object detection module relies on the outliers of frame localization. Thus, we define the dominant motion pattern as the static map, whereas the segments with high number of outliers initialize an object map. In other words, the objects are seen as a small subgroup of features from the input frames. However, static and object maps are just a matter of naming. Our algorithm would successfully work in the case, where the sensor is zoomed in to the object and the dominant motion pattern comes from the object. The key strength of our method is the multi-group registration procedure that considers the whole set of measurements and subsets of measurements for localization. It is also worth mentioning that the system can have difficulty detecting two different moving objects, if their motion is visible at the same frame. In this case, our system will initialize a single map for both objects and will fail to grow the object map. In the future, we would like to consider a sequence of frames for moving object detection instead of considering only consecutive frames. Another solution to this problem can be moving object detection in each object map in order to split multiple motion patterns.

One of the limitations of this work is convergence of the maps in case of a contamination from one of them to the other. This is due to the fact that the feature classification relies on the registration. Once some measurements are mistakenly added to the map, it might result in localization of incorrect regions. Our future work includes developing a pruning method that checks feature classification in past keyframes and corrects them in case of a misclassification. Also, the presented technique will have difficulty in texture-less areas and objects, since it is sparse feature-based.

This work focuses on rigidly moving objects in a scene. According to the presented approach, a moving object can be either continuously moving while being seen in the camera field of view or it can be seen in discrete time instances throughout the sequence, e.g., multiple instances appearing in far places in a scene. Our method can handle both situations. Due to our moving object assumption, the maps are independent from each other and they cannot share any geometric constraint. However, if the detected objects are stationary in some frames with respect to the static scene, the maps can be partially dependent to each other. This information can help improve accuracy in the bundle adjustment, which is another important future direction of this work.

%% file: 3DV_joint_reconstruction.bbl
\begin{thebibliography}{10}\itemsep=-1pt

\bibitem{CeresSolver}
S.~Agarwal, K.~Mierle, and Others.
\newblock Ceres solver.
\newblock \url{http://ceres-solver.org}.

\bibitem{Ambrus2017}
R.~Ambrus, N.~Bore, J.~Folkesson, and P.~Jensfelt.
\newblock Autonomous meshing, texturing and recognition of object models with a
  mobile robot.
\newblock In {\em The 2017 IEEE/RSJ International Conference on Intelligent
  Robots and Systems (IROS 2017)}, Oct. 2017.

\bibitem{Cansizoglu16IROS}
E.~Ataer-Cansizoglu and Y.~Taguchi.
\newblock Object detection and tracking in {RGB-D SLAM} via hierarchical
  feature grouping.
\newblock In {\em Proc. IEEE/RSJ Int'l Conf. Intelligent Robots and Systems
  (IROS)}, 2016.

\bibitem{Cansizoglu16ICRA}
E.~Ataer-Cansizoglu, Y.~Taguchi, and S.~Ramalingam.
\newblock Pinpoint {SLAM}: A hybrid of {2D} and {3D} simultaneous localization
  and mapping for {RGB-D} sensors.
\newblock In {\em Proc. IEEE Int'l Conf. Robotics and Automation (ICRA)}, 2016.

\bibitem{Cadena2016}
C.~Cadena, L.~Carlone, H.~Carrillo, Y.~Latif, D.~Scaramuzza, J.~Neira, I.~Reid,
  and J.~J. Leonard.
\newblock Past, present, and future of simultaneous localization and mapping:
  Toward the robust-perception age.
\newblock {\em IEEE Transactions on Robotics}, 32(6):1309--1332, Dec. 2016.

\bibitem{Choudhary2014}
S.~Choudhary, A.~J. Trevor, H.~I. Christensen, and F.~Dellaert.
\newblock {SLAM} with object discovery, modeling and mapping.
\newblock In {\em Proc. IEEE/RSJ Int'l Conf. Intelligent Robots and Systems
  (IROS)}, pages 1018--1025, Sept. 2014.

\bibitem{Civera2011}
J.~Civera, D.~G{\'a}lvez-L{\'o}pez, L.~Riazuelo, J.~D. Tard{\'o}s, and
  J.~Montiel.
\newblock Towards semantic {SLAM} using a monocular camera.
\newblock In {\em Proc. IEEE/RSJ Int'l Conf. Intelligent Robots and Systems
  (IROS)}, pages 1277--1284. IEEE, 2011.

\bibitem{Dame2013}
A.~Dame, V.~A. Prisacariu, C.~Y. Ren, and I.~Reid.
\newblock Dense reconstruction using {3D} object shape priors.
\newblock In {\em Proc. IEEE Conf. Computer Vision and Pattern Recognition
  (CVPR)}, pages 1288--1295, 2013.

\bibitem{Dharmasiri2016}
T.~Dharmasiri, V.~Lui, and T.~Drummond.
\newblock {MO-SLAM}: Multi object {SLAM} with run-time object discovery through
  duplicates.
\newblock In {\em Proc. IEEE/RSJ Int'l Conf. Intelligent Robots and Systems
  (IROS)}, pages 1214--1221, 2016.

\bibitem{Faulhammer2017}
T.~F{\"a}ulhammer, R.~Ambru{\c{s}}, C.~Burbridge, M.~Zillich, J.~Folkesson,
  N.~Hawes, P.~Jensfelt, and M.~Vincze.
\newblock Autonomous learning of object models on a mobile robot.
\newblock {\em IEEE Robotics and Automation Letters}, 2(1):26--33, 2017.

\bibitem{Feng14ICRA}
C.~Feng, Y.~Taguchi, and V.~R. Kamat.
\newblock Fast plane extraction in organized point clouds using agglomerative
  hierarchical clustering.
\newblock In {\em Proc. IEEE Int'l Conf. Robotics and Automation (ICRA)}, May
  2014.

\bibitem{Finman2013}
R.~Finman, T.~Whelan, M.~Kaess, and J.~J. Leonard.
\newblock Toward lifelong object segmentation from change detection in dense
  {RGB-D} maps.
\newblock In {\em European Conference on Mobile Robots (ECMR)}, pages 178--185,
  2013.

\bibitem{Fioraio2013}
N.~Fioraio and L.~D. Stefano.
\newblock Joint detection, tracking and mapping by semantic bundle adjustment.
\newblock In {\em Proc. IEEE Conf. Computer Vision and Pattern Recognition
  (CVPR)}, pages 1538--1545, 2013.

\bibitem{Jegou12PAMI}
H.~J\'egou, F.~Perronnin, M.~Douze, J.~S\'anchez, P.~P\'erez, and C.~Schmid.
\newblock Aggregating local image descriptors into compact codes.
\newblock {\em {IEEE} Trans. Pattern Anal. Mach. Intell.}, 34(9):1704--1716,
  Sept. 2012.

\bibitem{Jiang2016}
C.~Jiang, D.~P. Paudel, Y.~Fougerolle, D.~Fofi, and C.~Demonceaux.
\newblock Static-map and dynamic object reconstruction in outdoor scenes using
  {3-D} motion segmentation.
\newblock {\em IEEE Robotics and Automation Letters}, 1(1):324--331, Jan. 2016.

\bibitem{Keller2013}
M.~Keller, D.~Lefloch, M.~Lambers, S.~Izadi, T.~Weyrich, and A.~Kolb.
\newblock Real-time {3D} reconstruction in dynamic scenes using point-based
  fusion.
\newblock In {\em Proc. Int'l Conf. {3D} Vision (3DV)}, pages 1--8, June 2013.

\bibitem{Khoshelham2012Accuracy}
K.~Khoshelham and S.~O. Elberink.
\newblock Accuracy and resolution of {Kinect} depth data for indoor mapping
  applications.
\newblock {\em Sensors}, 12(2):1437--1454, 2012.

\bibitem{Lowe04IJCV}
D.~G. Lowe.
\newblock Distinctive image features from scale-invariant keypoints.
\newblock {\em Int'l J. Computer Vision}, 60(2):91--110, 2004.

\bibitem{Ma2014}
L.~Ma and G.~Sibley.
\newblock Unsupervised dense object discovery, detection, tracking and
  reconstruction.
\newblock In {\em Proc. European Conf. Computer Vision (ECCV)}, pages 80--95,
  2014.

\bibitem{Mustafa2015}
A.~Mustafa, H.~Kim, J.-Y. Guillemaut, and A.~Hilton.
\newblock General dynamic scene reconstruction from multiple view video.
\newblock In {\em Proc. IEEE Int'l Conf. Computer Vision (ICCV)}, pages
  900--908, 2015.

\bibitem{SalasMoreno13CVPR}
R.~F. Salas-Moreno, R.~A. Newcombe, H.~Strasdat, P.~H.~J. Kelly, and A.~J.
  Davison.
\newblock {SLAM++}: Simultaneous localisation and mapping at the level of
  objects.
\newblock In {\em Proc. IEEE Conf. Computer Vision and Pattern Recognition
  (CVPR)}, June 2013.

\bibitem{Shin2013}
Y.~M. Shin, M.~Cho, and K.~M. Lee.
\newblock Multi-object reconstruction from dynamic scenes: An object-centered
  approach.
\newblock {\em Computer Vision and Image Understanding}, 117(11):1575--1588,
  Nov. 2013.

\bibitem{Taguchi13ICRA}
Y.~Taguchi, Y.-D. Jian, S.~Ramalingam, and C.~Feng.
\newblock Point-plane {SLAM} for hand-held {3D} sensors.
\newblock In {\em Proc. IEEE Int'l Conf. Robotics and Automation (ICRA)}, pages
  5182--5189, May 2013.

\bibitem{Tan2013}
W.~Tan, H.~Liu, Z.~Dong, G.~Zhang, and H.~Bao.
\newblock Robust monocular {SLAM} in dynamic environments.
\newblock In {\em Proc. IEEE Int'l Symp. Mixed and Augmented Reality (ISMAR)},
  pages 209--218, 2013.

\bibitem{Wang2007}
C.-C. Wang, C.~Thorpe, S.~Thrun, M.~Hebert, and H.~Durrant-Whyte.
\newblock Simultaneous localization, mapping and moving object tracking.
\newblock {\em Int'l J. Robotics Research}, 26(9):889--916, Sept. 2007.

\bibitem{Yuheng2013}
C.~Yuheng~Ren, V.~Prisacariu, D.~Murray, and I.~Reid.
\newblock {STAR3D}: Simultaneous tracking and reconstruction of {3D} objects
  using {RGB-D} data.
\newblock In {\em Proc. IEEE Int'l Conf. Computer Vision (ICCV)}, pages
  1561--1568, 2013.

\end{thebibliography}
